\pdfoutput=1

\documentclass[11pt]{article}

\usepackage[final]{acl}

\usepackage{times}
\usepackage{latexsym}

\usepackage[T1]{fontenc}

\usepackage[utf8]{inputenc}

\usepackage{microtype}

\usepackage{inconsolata}

\usepackage{graphicx}

\usepackage{enumitem}%
\usepackage{tcolorbox}
\usepackage{helvet}
\usepackage{float}

\title{Enhancing Discoverability in Enterprise Conversational Systems with Proactive Question Suggestions}

\author{
  \textbf{Xiaobin Shen\textsuperscript{1}\thanks{Work done as an intern at Adobe Inc.}},
 \textbf{Daniel Lee\textsuperscript{2}},
 \textbf{Sumit Ranjan\textsuperscript{2}},\\
 \textbf{Sai Sree Harsha\textsuperscript{2}},
 \textbf{Pawan Sevak\textsuperscript{2}},
 \textbf{Yunyao Li\textsuperscript{2}}\\
 \textsuperscript{1}Carnegie Mellon University,
 \textsuperscript{2}Adobe
\\
 \small{
   \texttt{xiaobins@andrew.cmu.edu, dlee1@adobe.com, yunyaol@adobe.com} }
  }

\begin{document}
\maketitle

\begin{abstract}
Enterprise conversational AI systems\footnote{\url{https://www.gartner.com/reviews/market/enterprise-conversational-ai-platforms}} are becoming increasingly popular to assist users in completing daily tasks such as those in marketing and customer management. However, new users often struggle to ask effective questions, especially in emerging systems with unfamiliar or evolving capabilities. This paper proposes a framework to enhance question suggestions in conversational enterprise AI systems by generating proactive, context-aware questions that try to address immediate user needs while improving feature discoverability. Our approach combines periodic user intent analysis at the population level with chat session-based question generation. We evaluate the framework using real-world data from the AI Assistant for Adobe Experience Platform (AEP), demonstrating the improved usefulness and system discoverability of the AI Assistant.
\end{abstract}

\vspace{-0.3em}
\section{Introduction}\label{sec:intro}
\vspace{-0.5em}

The rise of large language models (LLMs) (\citealp{10.5555/3648699.3648939, touvron2023llama, achiam2023gpt, dubey2024llama}) and retrieval-augmented generation (RAG) \citep{chen-etal-2017-reading} techniques has expanded the capabilities of AI systems, enabling their application in specialized domains such as conversational AI systems. It is increasingly common for companies to integrate AI Assistants into their products, with the aim of automating interactions and enhancing user experience.
In this context, enterprise conversational AI systems are being adopted into platforms like marketing and customer management systems \citep{maharaj-etal-2024-evaluation} to assist users with structured tasks while improving overall platform usability. 
Enterprise conversational AI systems typically handle two types of tasks: addressing product knowledge (e.g., explaining concepts) and generating operational insights (e.g., querying customer data). This paper focuses on the first type, which helps users understand concepts and navigate the platform effectively.

Despite these advances, question-and-answer (QA) alone is often insufficient to meet user needs in complex enterprise environments. Users frequently struggle to know what to ask next after receiving an answer \citep{white2009exploratory, 10.1145/3411764.3445618}, particularly in emerging enterprise AI systems where new users are unfamiliar with the platform's capabilities, and the system itself is continually evolving. 
For instance, consider a new employee from a marketing team who asks, ``How is profile richness calculated?'' and receives a long response explaining the metrics used in Adobe Experience Platform (AEP). Although the answer provides a basic understanding of profile richness, the user may still be left uncertain about how to apply this knowledge in practice, what actions they should take next, or how it fits into the broader context of their work. This uncertainty exemplifies the difficulty of framing subsequent questions that fully leverage the platform's potential. %

Question suggestions can play a vital role in mitigating this gap by not only responding to user queries but also proactively guiding users \citep{10.1145/3613904.3642168} toward relevant questions they may not have considered. 
In the earlier example, suggestions like ``What are the implications of exceeding the Profile Richness entitlement?'' or ``What are the steps to monitor and manage Profile Richness effectively?'' can direct the user to a deeper understanding of how profile richness impacts their work and encourage exploration of related functionalities. In enterprise AI settings, where effective use of the system hinges on understanding its complex functionality, well-designed question suggestions are essential for addressing immediate user needs and improving the discoverability \citep{10.1145/3640543.3645201} of underutilized capabilities.

However, question suggestion in enterprise AI presents the challenges of data sparsity and noisiness. Enterprise systems often lack sufficient historical interaction data to support conventional machine learning models for query prediction. Moreover, users may pose spontaneous or ill-structured questions that do not conform to clear patterns, complicating question generation. Additionally, as enterprise AI assistants evolve with new capabilities, there is a growing gap between the system's full functionality and what users know about it. This discoverability problem (\citealp{doi:10.1080/07370024.2024.2364606}) can lead to reduced engagement and incomplete adoption of the platform.

In this paper, we propose a framework (Figure~\ref{fig:query-suggest-diagram}) to address these challenges by enhancing question suggestions in enterprise conversational AI systems. We apply this framework to real-world interactions from AI Assistant in Adobe Experience Platform (AEP), a leading customer data marketing platform where the AI assistant was introduced in 2024\footnote{\url{https://experienceleague.adobe.com/en/docs/experience-platform/ai-assistant/home} }. Our approach focuses on generating proactive, categorized question suggestions to enhance discoverability within the platform.
In summary, our contribution includes:
\begin{itemize}[noitemsep,topsep=0pt]
    \item Propose a framework for next-question generation in enterprise conversational AI systems, combining population-level periodical user intent analysis with chat-session-level question suggestion.
    \item Use LLMs to generate contextual, categorized question suggestions based on current queries, past interactions, and relevant documents.
    \item Conduct a human evaluation to assess the quality of these question suggestions using multiple criteria, including relatedness, validness, usefulness, diversity, and discoverability.
\end{itemize}
To the best of our knowledge, this is the first study to investigate the impact of question suggestion in a real-world enterprise conversational AI system deployed in a widely adopted platform.

\section{Related Work}
\vspace{-0.5em}
\subsection{Question Suggestion}
Traditional question suggestion techniques have been pivotal in enhancing user experience in search engines by predicting and recommending subsequent questions based on previous user interactions. These methods span from statistical approaches (e.g. \citealp{10.1145/2505515.2505661, 6816668, 10.1145/3020165.3022129, 10.1145/3077136.3080652}) to neural network methods (e.g. \citealp{10.1145/2806416.2806493, 10.1145/3132847.3133010, 10.1145/3209978.3210079, 10.1145/3397271.3401331, 10.1145/3470562}), all of which have significantly impacted large-scale web search applications.
Furthermore, some works attempt to diversify question suggestions by generating varied recommendations (e.g. \citealp{10.1145/2505515.2505661, 6816668}). This diversification is crucial in offering users broader and more relevant options.
These methods typically rely on large amounts of task-specific data to train models capable of making accurate predictions from scratch.

Advances in pre-trained LLMs and RAG have reduced reliance on task-specific data by leveraging pre-trained knowledge and retrieval, enhancing question suggestion capabilities. For example, \citet{10.1145/3589334.3645404} enrich the query suggestion process by integrating users' external domain-specific knowledge into LLMs, resulting in more contextually relevant recommendations. Similarly, \citet{bacciu2024generating} generate suggestions using LLMs without requiring direct access to user data, employing RAG-based on search logs. \citet{10.1145/3589334.3645365} extend question suggestion beyond textual input by incorporating image-based queries, enhancing both diversity and relevance through multi-agent reinforcement learning with LLMs.

In the context of emerging enterprise conversational systems \citep{maharaj-etal-2024-evaluation}, the challenge of effective question suggestion is compounded by the limited availability of high-quality training data. These systems often face constraints such as sparse historical interaction data, unpredictable user queries, and the absence of discernible query patterns. As a result, it becomes critical to design question suggestion strategies that effectively guide users despite the data limitation. Leveraging LLMs alongside efficient design choices can help overcome these challenges.

Moreover, the goal of question suggestions in enterprise conversation systems differs from those in general search engines. Not only do we want to recommend contextually relevant questions, but we also need to guide users toward exploring system capabilities and learning about new functionalities (more discussion in Section~\ref{sec:discover-literature}).

Existing methods often focus on general search contexts, overlooking the unique requirements of enterprise conversational systems. Our approach helps bridge this gap by considering both the immediate user interaction context and the specific intents and objectives within enterprise settings. %

\vspace{-0.3em}
\subsection{Discoverability in Virtual/AI Assistant} \label{sec:discover-literature}
\vspace{-0.3em}

Discoverability, as defined by \citet{norman2013design}, refers to the user's ability to ``\textit{figure out what actions are possible [with a system] and where and how to perform them}''. While this concept has been widely studied in the context of software and hardware interfaces, it is often overlooked in more complex systems such as AI-driven platforms and enterprise applications. 

Discoverability becomes increasingly important as systems grow more capable and feature-rich. Recent works (e.g., \citealp{doi:10.1080/07370024.2024.2364606}) emphasize that as systems become more complex, users often struggle to identify new or updated functionalities, leading to decreased engagement or adoption. Without proper guidance, users are overwhelmed by the growing feature set, which impedes their ability to leverage the platform fully.

Previous studies on discoverability focus primarily on desktop software (e.g. \citealp{10.1145/1622176.1622214, 10.1145/2470654.2466442, 10.1145/1970378.1970380}), smartphones applications (e.g. \citealp{10.1145/3240167.3240176}), and voice user interfaces (e.g. \citealp{10.1145/3027063.3053166, white2018skill, 10.1145/3405755.3406119}). Many of these works aim to suggest new or unused commands relevant to the user or context, thus improving the overall user experience and encouraging engagement with new functionalities \citep{10.1145/1622176.1622214, 10.1145/2470654.2466442}.

Recent studies have begun to explore the benefits of proactive interactions in conversational AI. For example, \citet{10.1145/3613904.3642168} demonstrate that proactive dialogues and timely suggestions can enhance user experience. Transitioning from user-directed dialogue to system-directed, such as providing follow-up questions or suggestions, can lead to more meaningful interactions and better user satisfaction.
\citet{10.1145/3640543.3645201} show that interactive conversational recommendations can lead to better discoverability of system functionalities. 
Users often struggle to articulate queries due to a lack of domain-specific language or well-defined informational goals \citep{10.1145/3411764.3445618}.

In enterprise conversational AI systems, discoverability remains underexplored, particularly in domains where users must engage with intricate systems in a business context (e.g. customer management). Users in these settings often have diverse roles and objectives, making it essential for the system to facilitate both immediate user engagement and long-term education about system capabilities. %

In summary, while prior work has addressed discoverability in various contexts, there is a gap specifically targeting enterprise conversational AI systems. Our approach aims to bridge this gap by designing a question suggestion strategy that not only guides users in their immediate tasks but also educates them about the system's capabilities, thereby fostering sustainable user engagement.

\section{Query Suggestion in an Enterprise Conversational AI System}\label{sec:method}
\vspace{-0.5em}

\begin{figure*}[!htbp]
\vspace{-0.3em}
  \includegraphics[width=1.0\linewidth]{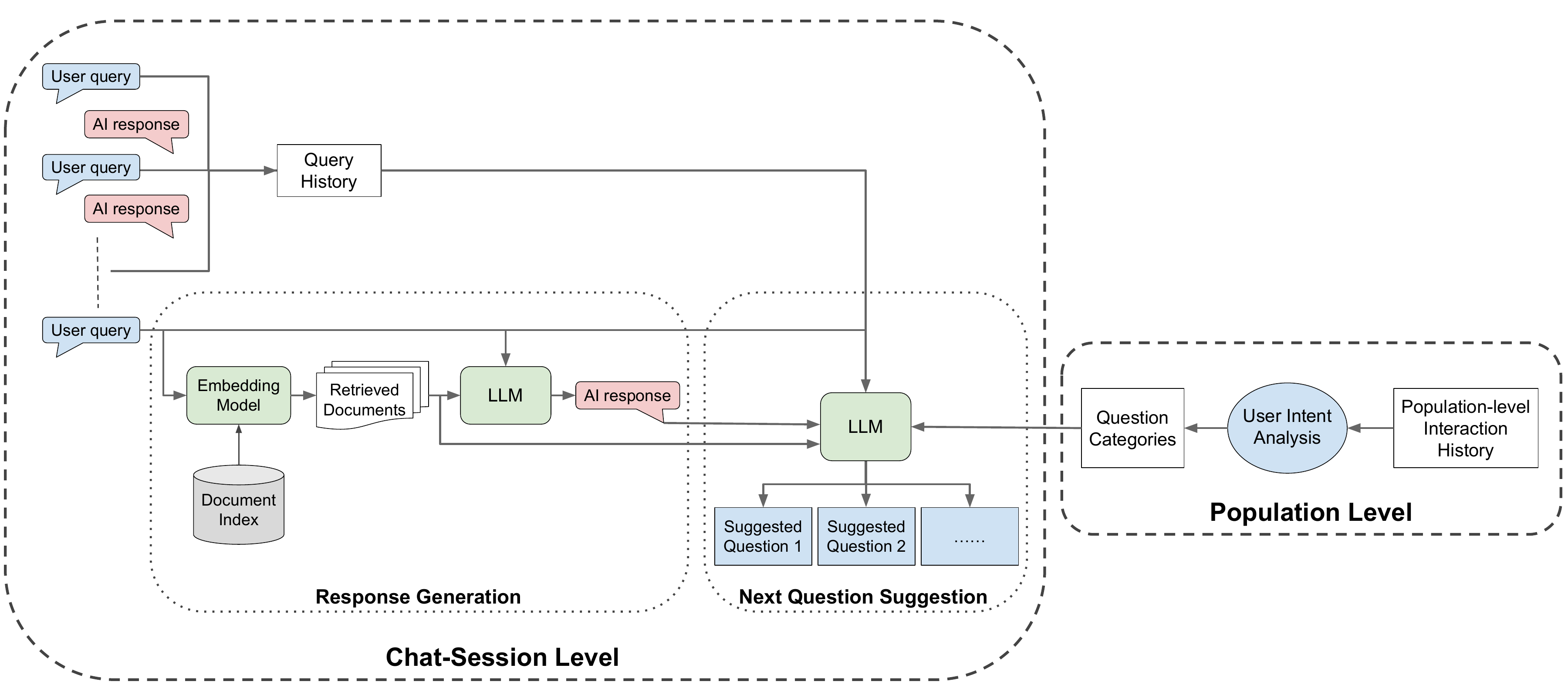} \centering
  \caption {Next Question Suggestion Framework in Enterprise Conversational AI Systems. The framework consists of two components: \textbf{population-level user intent analysis} (right), which generates question categories used for \textbf{next-question suggestion at the chat-session level} (middle). The response generation process (left) is included for completeness, as it precedes the question suggestion step and shares several inputs and outputs, such as the user query, retrieved documents, and AI response.
  \label{fig:query-suggest-diagram}}
  \vspace{-0.6em}
\end{figure*}

To address the challenges of data sparsity and noisiness in question suggestions for enterprise conversational AI systems, as well as to improve discoverability, we propose a simple yet efficient framework for generating next-question suggestions based on the user’s inquiry and the corresponding AI system response. Our framework leverages the interaction history within the same chat session as contextual information to generate relevant suggestions.

As shown in Figure~\ref{fig:query-suggest-diagram}, our framework consists of two key stages: user intent analysis and question generation. The user intent analysis is conducted periodically across the system’s entire user base to identify shifts in user behavior and emerging intents. 
The question generation stage, on the other hand, operates at the chat-session level, where it considers the user's interaction history within a session to appropriate relevant next questions.

\vspace{-0.3em}
\subsection{QA Interaction Data} 
\vspace{-0.3em}
In our setup, each chat session is represented by an interaction history specific to that session, denoted as $\mathcal{S}_i \hspace{-0.2cm} = \hspace{-0.2cm} \{(q_{i,1}, r_{i,1}, \mathcal{D}_{i,1}), ..., (q_{i,T}, r_{i,T}, \mathcal{D}_{i,T})\}$, where $q_{i,j}$ is the $j$-th question asked by the user in the current session, $r_{i,j}$ is the AI system’s response to $q_{i,j}$, $\mathcal{D}_{i,j}$ represents the set of documents retrieved to generate the response $r_{i,j}$, $T$ denotes the number of rounds of interaction in the session. We restrict the context to be only within the same chat session for a user, instead of all previous QA interactions, which can be lengthy and irrelevant.
Notice that we are recommending the next question, so at least one interaction is available at the time of question suggestion.

\vspace{-0.3em}
\subsection{Population Level User Intent Analysis}
\vspace{-0.3em}
Assume the aggregated interaction history $\mathcal{H} = \bigcup_{j=1}^{n} \mathcal{H}_i$ across all users' interaction sessions with the system, where $\mathcal{H}_i$ is the interaction history for chat sessions $j$ across multiple sessions. The user intent analysis (Figure~\ref{fig:query-suggest-diagram}, right) identifies common patterns and emerging user intents, which are then categorized into types $\mathcal{C} = \{c_k\}, k \in [K]$. This step is done to understand why users ask a question, conditional on the current contextual interaction, and is usually done externally by product people who really understand users. For example, when a user asks how to perform a certain task, they might become confused about some connected concepts mentioned in AI's response and then need to ask follow-up questions to further explore those topics.
This is crucial for an emerging enterprise conversational system where there is initially no clear understanding of why users ask their next questions, and user intents may evolve over time as new functionalities are introduced. This analysis helps the system identify broader trends and emerging needs within the user base, which can guide the types of questions generated during individual chat sessions. As noted in Section~\ref{sec:exp-result}, question categories such as \{\texttt{Exansion}, \texttt{Follow-up}\} can address specific types of queries that users may ask in response to the AI's answers. The generated categories are specifically tailored for guiding users to related but potentially unknown features, thus enhancing system discoverability. This user intent analysis phase can be conducted periodically as the system and user base expand, allowing new question categories to be introduced as needed.

\vspace{-0.5em}
\subsection{Chat-Session Level Question Suggestion}
\vspace{-0.3em}
The question generation phase (Figure~\ref{fig:query-suggest-diagram}, left) is conducted during individual chat sessions, using the current interaction history $\mathcal{S}_i$ and the most recent user intents identified during the periodic user intent analysis. The inputs for this phase include: (1) the current query $q_{i,t}$, the most recent question in the session; (2) the current response $r_{i,t}$, the AI’s response to the current query; (3) the questions previously asked by the user in the same chat session $\{q_{i, 1}, ..., q_{i, t-1}\}$, and could be empty; (4) the relevant documents $\mathcal{D}_{i,t}$: the documents retrieved during response generation; and (5) question categories $\mathcal{C}$ from user intent analysis.

Leveraging these inputs, the LLM generates suggested questions that align with the predefined intent categories (e.g. \texttt{Exansion}, \texttt{Follow-up}). The LLM's ability to understand context and generate diverse question types ensures that even when the session history is limited, the suggestions remain contextually relevant and proactive. Furthermore, by incorporating relevant documents $\mathcal{D}_{i,t}$, the LLM can enhance the discoverability of features and information within the system, as the suggestions will include more precise and domain-specific content.

By leveraging both user intent analysis at the population level and real-time session data, our framework aims to provide proactive question suggestions that guide the user’s interaction with the system. As new functionalities are introduced and user behavior shifts, the intent analysis can be updated, allowing the system to continuously refine the types of follow-up questions it generates. 
Our framework also ensures that even with limited session history, the system can produce suggestions that anticipate the user’s needs, encourage exploration of new features, and ultimately improve user satisfaction and discoverability within the platform.

\section{Evaluation} \label{sec:exp-result}
\vspace{-0.3em}
Enhancing discoverability in enterprise conversational systems is a novel and challenging topic (as mentioned in Sections~\ref{sec:intro} and \ref{sec:discover-literature}), without standard datasets or evaluation metrics. We curated our evaluation datasets from randomly selected 250 interactions between real users and the AI Assistant in AEP and, following the approach of \citet{10.1145/3589334.3645404}, employed human evaluation using carefully designed metrics to ensure a robust assessment of our framework.

\vspace{-0.1em}
\subsection{User Intent Analysis} \label{sec:user-intent-result}
\vspace{-0.2em}
To better understand user intent at the population level, we conducted an analysis of users' next-question patterns, conditioned on the current question, the AI Assistant’s response, and up to five prior questions within the same session. 

The findings, summarized in Table~\ref{tab:user-intent-analysis}, reveal that over 35\% of user queries were not related to previous interactions within the session, highlighting the challenges of modeling query patterns for enterprise AI systems with limited data. Additionally, many users tend to ask expansion questions (30\%) or follow-up questions (11\%) related to the AI Assistant’s response. These insights motivated us to define two primary categories for question suggestions: Expansion and Follow-Up, to better capture the diverse user intents. Specifically,
\begin{itemize} [noitemsep,topsep=2pt]
    \item \texttt{Expansion}: Questions that expand on current topics with related concepts or details.%
    \item \texttt{Follow-Up}: Questions that follow up on the AI's response, especially when there are multiple steps involved in the response to achieve the user's goal.%
\end{itemize}

\vspace{-0.2em}
\subsection{Human Evaluation}
\vspace{-0.2em}
We used GPT-3.5 \cite{openai2023gpt35} to generate both the baseline and enhanced question suggestions, ensuring a fair comparison. The baseline suggestions were generated immediately after the response to the user's question using a combined prompt for response and question suggestion generation, without considering the interaction history or questions categories. Our enhanced approach, on the other hand, uses a separate prompt for question generation by explicitly including the earlier generated response, pre-defined question categories, and interaction history (see Appendix~\ref{app:prompt-template} for the prompt template) to guide the LLM in generating suggestions tailored to the user's past interactions. The same document retrieval methods \cite{10.5555/3495724.3496517} were applied to both methods.
To see an example of question suggestion using a real snapshot of user interaction using our solution, please see Figure~\ref{fig:query-suggest-example}.

\begin{table}[t]
    \centering
    \begin{tabular}{lc}
        \hline
        \textbf{Category} & \textbf{Proportion} \\ \hline
        Unrelated things & 36\% \\
        Expansion on related topics & 30\% \\
        Follow-up of the response & 11\% \\
        Others & 23\% \\ \hline
    \end{tabular}
    \caption{User Intents based on Next Question Patterns.}
    \label{tab:user-intent-analysis}
    \vspace{-0.5em}
\end{table}

For human evaluation, we employed a pairwise comparison methodology, asking ten annotators to compare the question suggestions generated by the baseline and our framework based on five criteria: \textit{Relatedness}, \textit{Validity}, \textit{Usefulness}, \textit{Diversity}, and \textit{Discoverability}. Pairwise comparisons were chosen over independent ratings to encourage relative judgments, which tend to be more reliable \cite{stewart2005absolute}. 
Detailed definitions of these criteria and the annotation process are provided in Appendix~\ref{app:annotation-ui}. As for why we use the above five criteria,
\begin{itemize}[noitemsep,topsep=0pt]
    \item Relatedness, validity, and usefulness are adapted from \citet{10.1145/3589334.3645404}, considered as the standard for evaluating query suggestions.
    \item Diversity is added to ensure that the suggested questions cover a range of topics and are distinct from each other, especially since user intents can be diverse \cite{pradeep2024convkgyarnspinningconfigurablescalable}.
    \item Discoverability is included as a novel and key criterion to evaluate how well the suggestions guide users in exploring new features and information within the system.
\end{itemize}

\vspace{-0.2em}
\subsection{Results}
\vspace{-0.2em}

\begin{table*}[!htbp]
\vspace{-0.2em}
    \centering 
    \begin{tabular}{lccccc}
    \hline
    & Equally Good & Baseline Better & Ours Better & Both Bad & \\
    \hline
Relatedness     &       0.464 &            0.215 &        0.297 &     0.022 \\
Validness       &       0.685 &            0.134 &        0.167 &     0.009 \\
Usefulness      &       0.351 &            0.278 &        0.354 &     0.015 \\
Diversity       &       0.620 &            0.171 &        0.197 &     0.009 \\
Discoverability &       0.401 &            0.232 &        0.334 &     0.030 \\
    \hline
    \end{tabular}
    \caption{\label{tab:humen-eval-result}
    Human evaluation results show the proportion of annotations for each psychometric schema. }
    \vspace{-0.7em}
\end{table*}

The results of the human evaluation, summarized in Table~\ref{tab:humen-eval-result}, show that our framework outperformed the baseline across all five criteria. 
It is worth noting that many responses were rated “Equally Good,” particularly for validity (68.5\%) and relatedness (46.4\%), reflecting the LLM's ability to produce valid, contextually appropriate questions with a well-defined prompt.
The differences were more pronounced for usefulness and discoverability. Our framework achieved 35.4\% preference (versus 27.8\% for the baseline) for usefulness and 33.4\% preference (versus 23.2\%) for discoverability. This indicates that our method was perceived as more effective at guiding users and helping them uncover new functionalities within the system. This result is particularly relevant for emerging enterprise systems, where the goal is not only to generate contextually relevant questions but also to support user exploration and learning.

\noindent\textbf{Annotator Role-Based Insights}. We analyzed the evaluation results by annotator roles, differentiating between engineers and product ones, to represent the two typical types of actual users of AEP, e.g. operational engineers versus marketers (see Appendix~\ref{app:eval-result-by-annotator} for details). 
In Table~\ref{tab:eval-by-annotator}, 
product annotator P1 showed a strong preference for our approach over the baseline, especially in relatedness, usefulness, and discoverability. P2 also showed slight favor towards our approach (except diversity), although being more conservative. This suggests that our approach aligns more closely with product-focused users who prioritize relevance and utility. Conversely, engineer-focused annotator E2 rated the baseline higher in \textit{usefulness}, with "Baseline Better" ratings of 50.0\%, possibly reflecting their preference for immediate technical relevance and precision. However, engineers also recognized improvements in our approach for \textit{diversity} and \textit{discoverability}, indicating it helps explore new features from a technical standpoint. These differences highlight how perceptions of what makes a suggestion "useful" or "discoverable", despite the provided definitions, can vary based on the evaluator's role, emphasizing the importance of tailoring the suggestion to accommodate diverse needs.

\vspace{-0.2em}
\subsection{General Insights}
\vspace{-0.2em}

Our study highlights the inherent challenges of data sparsity and noisiness in user interaction logs for enterprise AI systems, as shown in Section~\ref{sec:user-intent-result}. Attempting to train a question suggestion model from scratch or fine-tuning a pre-trained language model is impractical under these conditions. Instead, leveraging LLMs with RAG is an effective solution for generating question suggestions in a context where platform functionality is complex and evolving, and user intents are diverse, as it is a proven trend in practice.

The findings also emphasize the need for personalization over a one-size-fits-all solution, aligning with existing research (e.g. \citealp{10.1145/3589334.3645404}) that demonstrates the variability in users’ intents. Additionally, the varying perceptions of different criteria of question suggestions among different annotators highlight the importance of diverse evaluation criteria. This variability suggests that multiple viewpoints should be considered when assessing the effectiveness of question suggestion systems, especially in enterprise environments where user expectations are varied and complex.

\vspace{-0.3em}
\section{Conclusion}
\label{sec:conclude}
\vspace{-0.3em}
In this paper, we presented a framework for enhancing question suggestions in enterprise conversational AI systems, balancing contextual relevance with improved discoverability to help users navigate complex platforms and explore underutilized features. Our evaluation demonstrates the framework’s effectiveness in addressing the unique challenges of enterprise AI, emphasizing the importance of guiding user exploration alongside immediate query responses. Our work reinforces the need for adaptable and proactive question suggestion frameworks that can accommodate evolving user behaviors and system functionalities, ultimately enhancing user satisfaction and discoverability within enterprise platforms.

For future work, we aim to evaluate the impact of improved question suggestions on user engagement metrics in a production environment, such as click-through rates and feature exploration, to understand the long-term benefits of our approach on user behavior and system utilization. Moreover, developing more fine-grained and personalized categories based on user role, the familiarity with the platform (e.g., whether the user is a novice or experienced), would further enhance discoverability and long-term engagement.

\section*{Acknowledgments}
We would like to express our gratitude to Justin Coglitore, Sally Fang, Shaista Hussain, 
Shun Jiang, 
Kai Lau, Dan Luo, Akash Maharaj, Rakesh R Menon, Yuan Tian, Richard Yang (ordered alphabetically), for their valuable feedback.

\bibliography{our}

\appendix

\section{Appendix}
\label{sec:appendix}

\subsection{Prompt Template} \label{app:prompt-template}
See Figure~\ref{fig:prompts} for the prompt template for question suggestions.

\begin{figure*} [!htbp]
\centering
{\sffamily \small  %
    \begin{tcolorbox}[colback=blue!5!white, colframe=blue!75!black, title=Prompts Template for Contextual and Categorized Question Suggestion, width=\textwidth]
    
    Assume the identity of an Adobe Experience Platform (AEP) Specialist. Your task is to suggest questions for users regarding Adobe Experience Platform (AEP).

    \vspace{3pt}
    \textbf{Context}\\
    The user has asked a question about Adobe Experience Platform (AEP) and received an answer from an AI assistant. You have access to some documents that may help you generate your response. These documents may be incomplete or irrelevant: \texttt{\{documents\}}\\
    You also have access to earlier queries asked by the users, arranged in chronological order. Use them only when necessary. This list may be empty:  \texttt{\{query\_history\}}

    \vspace{3pt}
    \textbf{Goal}\\
    Brainstorm and suggest follow-up questions of different categories to help the user better explore AEP and its capabilities. Provide additional value to the user's understanding of AEP, based on the user's query and the AI assistant's response. 

    \vspace{3pt}
    \textbf{Categories for Follow-Up Questions}\\
    Generate at least one follow-up question for each of the first four categories below. Include questions from the "Other" category if necessary.\\
    1. Expansion: These questions should expand on the current topics with related concepts or details. Aim to introduce new but related information that the user might find useful.\\
    2. Follow-up: These questions should follow up on the AI assistant's response, especially when there are multiple steps involved in the response to achieve the user's goal. They can also address any subsequent actions or considerations the user might have.\\
    3. Other: Any other relevant question that does not fit into the above categories but still provides additional value to the user's understanding of AEP.

    \vspace{3pt}
    \textbf{Specifications for Suggested Questions}\\
    1. Concise: Questions must be short and simple. \\
    2. Length: Each question must be in 8 to 15 words. \\
    3. Categorization: End each suggested question with a parenthesis indicating the category of the question \\
    4. ...

    \vspace{5pt}
    Each of the examples below may only include query suggestions for one of the categories just for illustration purposes. You should aim to provide at least one question for each of four categories in your response. 

    \vspace{5pt}
    \textit{\#\#\# START EXAMPLES} \\
    \textit{EXAMPLE 1:} \\
    \textit{<CURRENT QUERY>}: What is event forwarding and the configuration's steps for it?\\
    \textit{<AI ASSISTANT RESPONSE>}: 
    Event forwarding in Adobe Experience Platform allows you to send collected event data to a destination for server-side processing...\\
    \textit{<QUERY SUGGESTIONS>}: \\ 
    How to install and configure the Adobe Experience Platform Web SDK? (Expansion) \\
    How can I do this Build and install a library to implement the rule on my website? (Follow-up)
    
    \vspace{3pt}
    \textit{EXAMPLE 2:}
    ...\\
    \textit{\#\#\# END EXAMPLES}
    
    \vspace{5pt}
    Now go ahead and generate question suggestions to the current user question. \\
    \textit{<CURRENT QUERY>}: \texttt{\{query\}} \\ 
    \textit{<AI ASSISTANT RESPONSE>}: \texttt{\{response\}} \\
    \textit{<QUERY SUGGESTIONS>}:
    \end{tcolorbox}
} %
\caption{Prompt template used for generating contextual and categorized question suggestions. Certain details have been omitted due to space constraints or business confidentiality.}
\label{fig:prompts}
\end{figure*}

\subsection{Question Suggestion Example}

\begin{figure}[ht]
  \includegraphics[width=\linewidth]{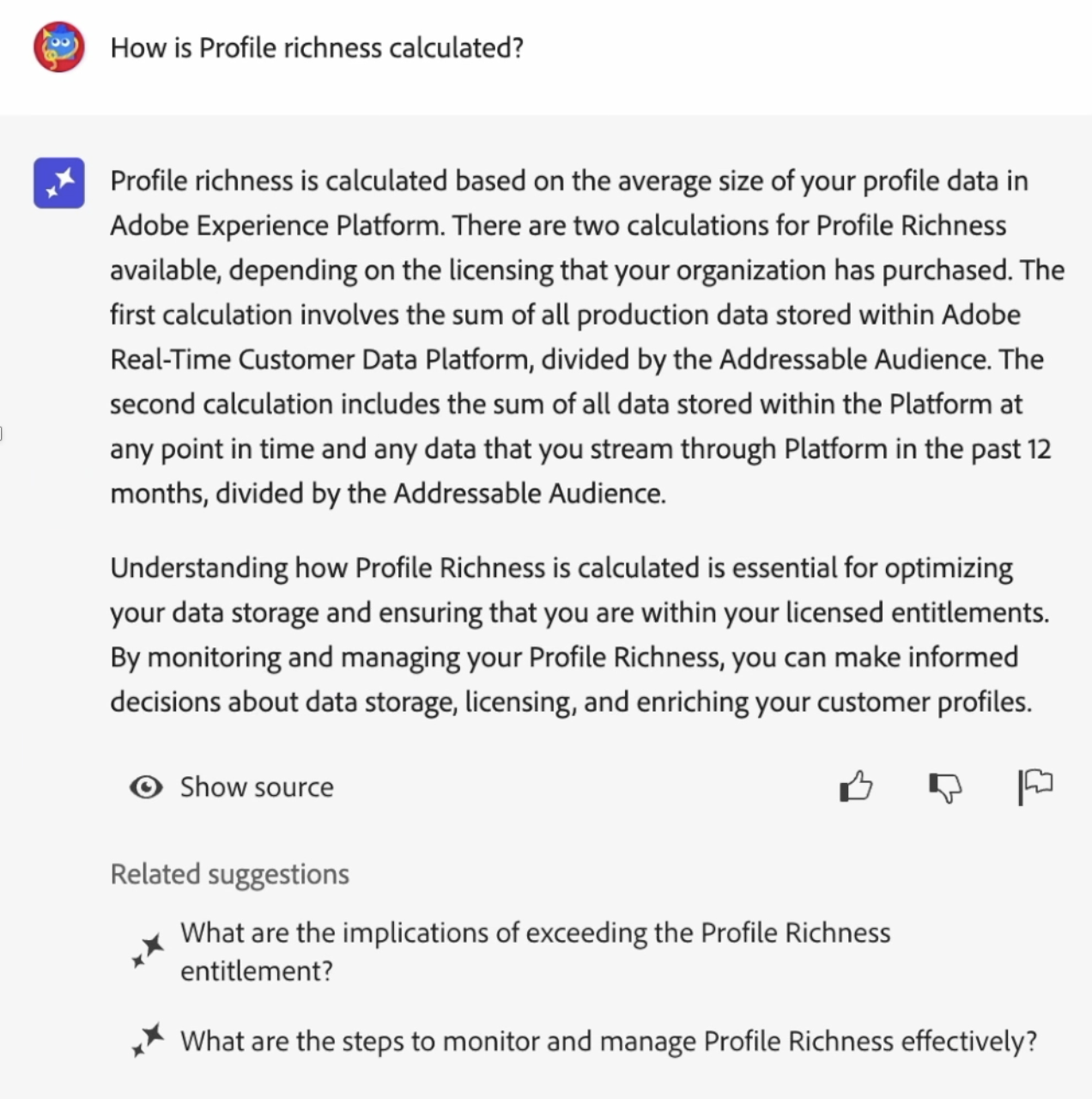}
  \caption {Example of question suggestions in AEP’s AI Assistant. When the user asks a question, the AI Assistant first provides a response to the question, followed by two suggested questions of different categories, with one expanding the related concepts mentioned in the question and response, and one suggesting potential future steps that can be taken to better utilize the platform functionalities.}
  \label{fig:query-suggest-example}
\end{figure}

Figure~\ref{fig:query-suggest-example} shows an example of question suggestions generated by AEP's AI Assistant. After the user asks, "How is profile richness calculated?", the AI Assistant provides a detailed response and then suggests two follow-up questions: one from the \texttt{Expansion} category ("What are the implications of exceeding the Profile Richness
entitlement?") and one from the \texttt{Follow-Up} category ("What are the steps to monitor and manage Profile Richness effectively?"). This example illustrates how question suggestions can help guide the user to explore related topics and understand the broader implications of the information provided.

\subsection{Human Evaluation Details} \label{app:annotation-ui}

The human evaluation was conducted based on five criteria, which are defined as follows:
\begin{itemize} [noitemsep,topsep=0pt]
    \item Relatedness: Measures whether the suggested queries are relevant to both the user's current query and the AI Assistant’s response.
    \item Validity: Assesses whether the suggested queries are valid questions within the context of Adobe Experience Platform (AEP) and, if possible to determine, whether they are answerable.
    \item Usefulness: Indicates how likely a user would be to select at least one of the suggested queries as their next question in the session.
    \item Diversity: Evaluates whether the suggested queries are distinct from one another and cover a broad range of topics.
    \item Discoverability: Determines whether the suggested queries help the user discover new features, capabilities, or information within AEP.
\end{itemize}

Each annotation was made considering the complete interaction context, which included the user's current query, the AI Assistant's response, and up to five previous queries within the same chat session. Annotators with familiarity in using Adobe Experience Platform (AEP) were recruited to perform the evaluation. To prevent bias, annotators were not informed which set of suggestions was generated by our method versus the baseline.

For each of the five criteria, annotators compared two sets of question suggestions (referred to as S1 and S2) and chose one of the following options:

\begin{itemize} [noitemsep,topsep=0pt]
    \item S1 is better than S2
    \item S1 and S2 are equally good
    \item S2 is better than S1
    \item None of them satisfy the criterion
\end{itemize}
The presentation order of S1 and S2 was randomized, meaning S1 could be either the baseline or our approach, to eliminate potential ordering bias. We briefed annotators before the evaluation to ensure accurate understanding of the evaluation criteria and response options.

\begin{figure*}[t]
    \centering
    \includegraphics[width=0.79\linewidth]{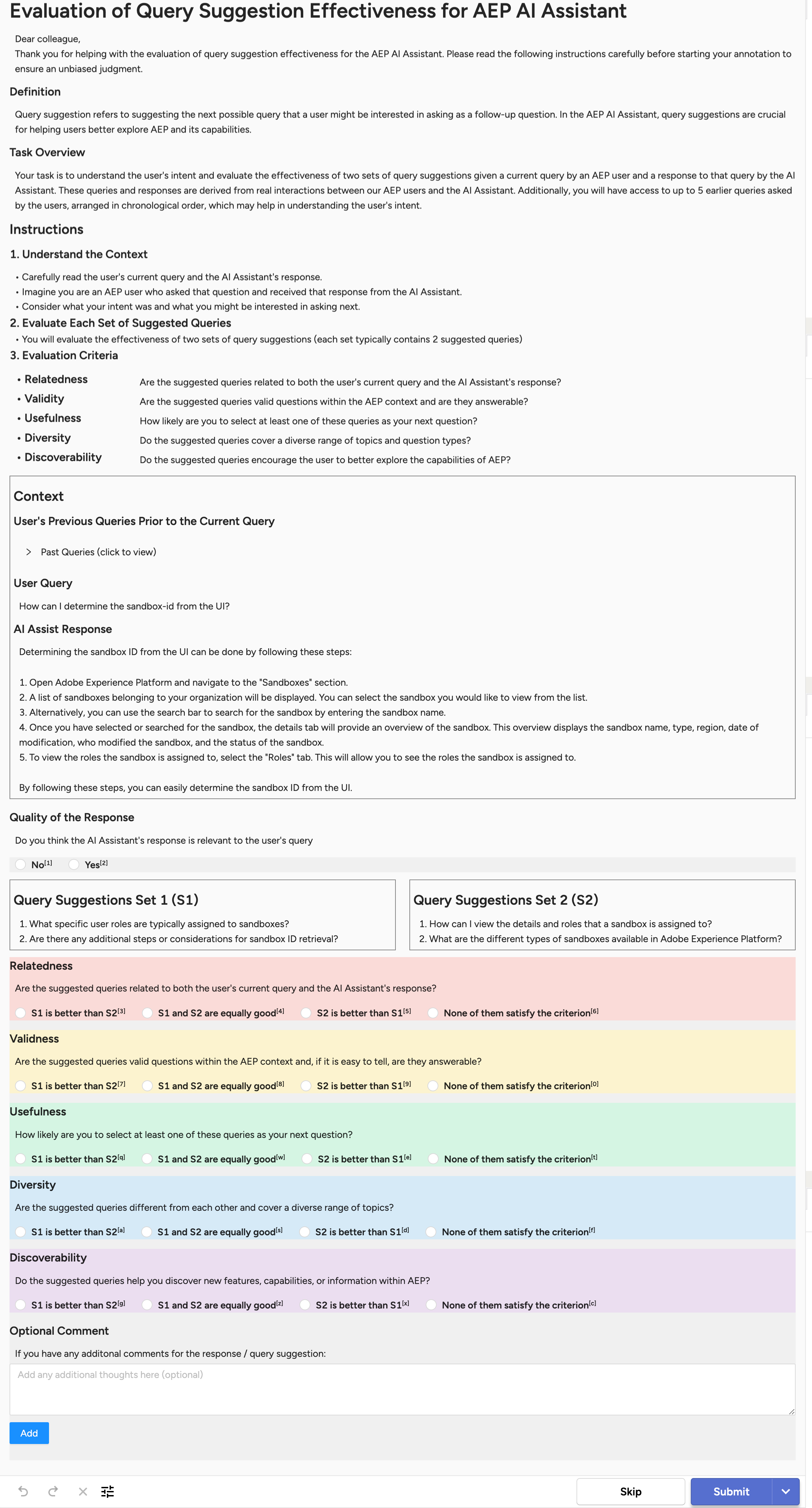}
    \caption {Human Evaluation Interface. Annotators used this interface to compare the quality of question suggestions based on the defined criteria.}
    \label{fig:human-eval-ui}
\end{figure*}

The annotation interface used for the evaluation is shown in  Figure~\ref{fig:human-eval-ui}. This interface provided annotators with the necessary context information and allowed them to select their ratings based on the comparison criteria.

\subsection{Human Evaluation Results by Annotators} \label{app:eval-result-by-annotator}

We display the detailed results from individual annotators, both from the product and engineer role. The distinguish between engineer and product roles is a genuine representation of the two typical types of users of the AEP. The engineer role, for example, could be someone who is responsible for customer data management and modeling; while the product role could be someone who set and trigger personalized campaigns.
The proportion of evaluations for each criterion was calculated for each annotator, and the results for all four annotators are presented in Table~\ref{tab:eval-by-annotator}. We only display four of the ten annotators for conciseness and the selected subset effectively captures the key trends and variations observed across the full group.

\begin{table*}%
    \centering 
    \begin{tabular}{lcccccc}
    \hline
         & Equally Good & Baseline Better & Ours Better & Both Bad & Annotator ID \\ \hline
        Relatedness & 0.517 & 0.183 & 0.233 & 0.067 & E1 \\
        Relatedness & 0.375 & 0.313 & 0.281 & 0.000 & E2 \\
        Relatedness & 0.206 & 0.216 & 0.578 & 0.000 & P1 \\
        Relatedness & 0.701 & 0.119 & 0.157 & 0.022 & P2 \\ \hline
        Validness & 0.800 & 0.100 & 0.067 & 0.017 & E1 \\ 
        Validness & 0.750 & 0.094 & 0.125 & 0.000 & E2 \\ 
        Validness & 0.588 & 0.108 & 0.304 & 0.000 & P1 \\ 
        Validness & 0.709 & 0.127 & 0.149 & 0.015 & P2 \\ \hline
        Usefulness & 0.133 & 0.400 & 0.433 & 0.033 & E1 \\ 
        Usefulness & 0.063 & 0.500 & 0.406 & 0.000 & E2 \\ 
        Usefulness & 0.206 & 0.225 & 0.569 & 0.000 & P1 \\ 
        Usefulness & 0.672 & 0.157 & 0.157 & 0.015 & P2 \\ \hline
        Diversity & 0.483 & 0.183 & 0.317 & 0.017 & E1 \\ 
        Diversity & 0.281 & 0.250 & 0.406 & 0.031 & E2 \\ 
        Diversity & 0.794 & 0.088 & 0.118 & 0.000 & P1 \\ 
        Diversity & 0.746 & 0.134 & 0.112 & 0.007 & P2 \\ \hline
        Discoverability & 0.133 & 0.250 & 0.617 & 0.000 & E1 \\ 
        Discoverability & 0.250 & 0.344 & 0.375 & 0.000 & E2 \\ 
        Discoverability & 0.225 & 0.216 & 0.559 & 0.000 & P1 \\ 
        Discoverability & 0.679 & 0.127 & 0.172 & 0.022 & P2 \\ \hline
    \end{tabular}
    \caption{Human evaluation results by individual annotators for each criterion. The columns indicate the proportion of annotations. The rows under each criterion correspond to different annotators: E (Engineer role) and P (Product role)}\label{tab:eval-by-annotator}
\end{table*}

\end{document}